\documentclass{article}

\usepackage[preprint]{iaseai26}

\usepackage[utf8]{inputenc} %
\usepackage[T1]{fontenc}    %
\usepackage{hyperref}       %
\usepackage{url}            %
\usepackage{booktabs}       %
\usepackage{amsfonts}       %
\usepackage{nicefrac}       %
\usepackage{microtype}      %

\usepackage{graphicx}

\usepackage{tikz}
\usetikzlibrary{calc}
\usetikzlibrary{decorations.pathreplacing}

\usepackage{booktabs}
\usepackage{colortbl}
\usepackage{makecell}
\usepackage{pifont}
\usepackage{amsmath}

\definecolor{good}{HTML}{CBF3CB} %
\definecolor{bad}{HTML}{FAD4D4} %
\definecolor{uncertain}{HTML}{FFF3B0} %

\newif\ifpost
\postfalse  %

\newcommand{\safe}{\cellcolor{good}\ding{51}}
\newcommand{\risk}{\cellcolor{bad}\ding{55}}
\newcommand{\unknown}{\cellcolor{uncertain}\textbf{?}}
\newcommand{\pipel}{pretraining$\rightarrow$SFT$\rightarrow$RLHF}

\title{AI Alignment Strategies from a Risk Perspective: Independent Safety Mechanisms or Shared Failures?}

\author{
\textbf{Leonard Dung}\textsuperscript{1} \quad
\textbf{Florian Mai}\textsuperscript{2,3} \\
\textsuperscript{1}Ruhr-Universität Bochum \\
\textsuperscript{2}Rheinische Friedrich-Wilhelms-Universität Bonn \\
\textsuperscript{3}Lamarr Institute for Machine Learning and Artificial Intelligence \\
\texttt{leonard.dung@ruhr-uni-bochum.de} \quad \texttt{fmai@uni-bonn.de}
}

\begin{document}

\maketitle

\begin{abstract}
AI alignment research aims to develop techniques to ensure that AI systems do not cause harm.
However, every alignment technique has \emph{failure modes}, which are conditions in which there is a non-negligible chance that the technique fails to provide safety.
As a strategy for risk mitigation, the AI safety community has increasingly adopted a \emph{defense-in-depth} framework:
Conceding that there is no single technique which guarantees safety, defense-in-depth consists in having multiple redundant protections against safety failure, such that safety can be maintained even if some protections fail.
However, the success of defense-in-depth depends on how (un)correlated failure modes are across alignment techniques. 
For example, if all techniques had the exact same failure modes, the defense-in-depth approach would provide no additional protection at all.
In this paper, we analyze 7 representative alignment techniques and 7 failure modes to understand the extent to which they overlap.
We then discuss our results' implications for understanding the current level of risk and how to prioritize AI alignment research in the future.

\end{abstract}

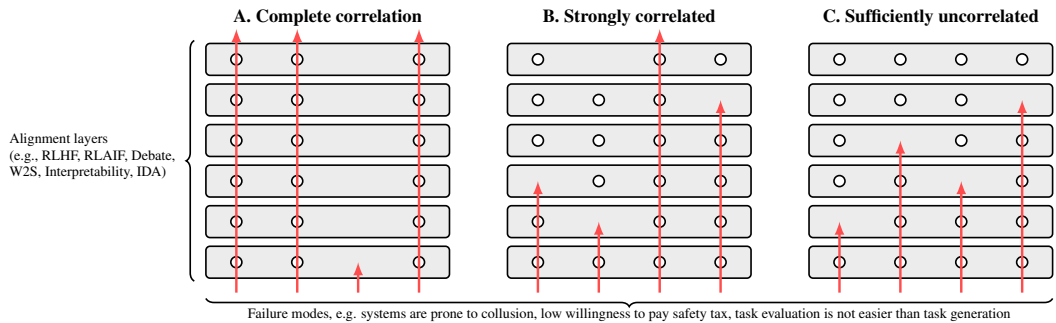
\begin{figure}[h]
\centering
\resizebox{\columnwidth}{!}{%
\begin{tikzpicture}[x=1cm,y=1cm,>=latex]
\tikzset{
  slice/.style={draw,rounded corners=2pt,fill=gray!15,thick},
  hole/.style={fill=white,draw=black,thick},
  title/.style={font=\bfseries\footnotesize},
  label/.style={font=\scriptsize,align=left},
  note/.style={draw,rounded corners=2pt,fill=gray!10,inner sep=6pt}
}
\pgfmathsetmacro{\W}{4.2}      %
\pgfmathsetmacro{\H}{0.55}     %
\pgfmathsetmacro{\G}{0.15}     %
\pgfmathsetmacro{\L}{6}        %
\pgfmathsetmacro{\layersep}{\H+\G}
\pgfmathsetmacro{\panelgap}{1.0}   %

\pgfmathsetmacro{\CxA}{0}
\pgfmathsetmacro{\CxB}{\CxA + \W + \panelgap}
\pgfmathsetmacro{\CxC}{\CxB + \W + \panelgap}
\pgfmathsetmacro{\XA}{\CxA - 0.5*\W}
\pgfmathsetmacro{\XB}{\CxB - 0.5*\W}
\pgfmathsetmacro{\XC}{\CxC - 0.5*\W}

\pgfmathsetmacro{\yTitle}{\L*\layersep + 0.3}        %
\pgfmathsetmacro{\yLo}{-0.25}                         %
\pgfmathsetmacro{\yHi}{\L*\layersep - \G + 0.25}      %

\node[title] at ({\CxA}, {\yTitle}) {A. Complete correlation};

\foreach \i in {0,...,5} {
  \pgfmathsetmacro{\y}{\i*\layersep}
  \draw[slice] ({\XA},{\y}) rectangle ({\XA+\W},{\y+\H});
  \foreach \j in {1,2,4} {
    \pgfmathsetmacro{\xj}{\XA + (\j-0.5)*\W/4}
    \fill[hole] ({\xj},{\y + 0.5*\H}) circle[radius=0.095];
  }
}
\foreach \j in {1,2,4} {
  \pgfmathsetmacro{\xj}{\XA + (\j-0.5)*\W/4}
  \draw[->,very thick,red!70] ({\xj},{\yLo}) -- ({\xj},{\yHi});
}
\pgfmathsetmacro{\xAthree}{\XA + (3-0.5)*\W/4}
\pgfmathsetmacro{\yStopAthree}{0*\layersep + 0.5*\H} %
\draw[->,very thick,red!70] ({\xAthree},{\yLo}) -- ({\xAthree},{\yStopAthree});

\draw[decorate,decoration={brace,amplitude=4pt},thick]
  ({\XA-0.2},{-0.05}) -- ({\XA-0.2},{\L*\layersep-\G+0.05});
\node[label,anchor=east] at ({\XA-0.4},{0.5*\L*\layersep}) {Alignment layers\\(e.g., RLHF, RLAIF, Debate,\\W2S, Interpretability, IDA)};

\node[title] at ({\CxB}, {\yTitle}) {B. Strongly correlated};

\foreach \i in {0,...,5} {
  \pgfmathsetmacro{\y}{\i*\layersep}
  \draw[slice] ({\XB},{\y}) rectangle ({\XB+\W},{\y+\H});
}
\pgfmathsetmacro{\xBone}{\XB + (1-0.5)*\W/4}
\foreach \i in {0,1,3,4,5} { \pgfmathsetmacro{\y}{\i*\layersep}
  \fill[hole] ({\xBone},{\y + 0.5*\H}) circle[radius=0.095]; }
\pgfmathsetmacro{\yStopBone}{2*\layersep + 0.5*\H}
\draw[->,very thick,red!70] ({\xBone},{\yLo}) -- ({\xBone},{\yStopBone});
\pgfmathsetmacro{\xBtwo}{\XB + (2-0.5)*\W/4}
\foreach \i in {0,2,3,4} { \pgfmathsetmacro{\y}{\i*\layersep}
  \fill[hole] ({\xBtwo},{\y + 0.5*\H}) circle[radius=0.095]; }
\pgfmathsetmacro{\yStopBtwo}{1*\layersep + 0.5*\H}
\draw[->,very thick,red!70] ({\xBtwo},{\yLo}) -- ({\xBtwo},{\yStopBtwo});
\pgfmathsetmacro{\xBthree}{\XB + (3-0.5)*\W/4}
\foreach \i in {0,1,2,3,4,5} { \pgfmathsetmacro{\y}{\i*\layersep}
  \fill[hole] ({\xBthree},{\y + 0.5*\H}) circle[radius=0.095]; }
\draw[->,very thick,red!70] ({\xBthree},{\yLo}) -- ({\xBthree},{\yHi});
\pgfmathsetmacro{\xBfour}{\XB + (4-0.5)*\W/4}
\foreach \i in {0,1,2,3,5} { \pgfmathsetmacro{\y}{\i*\layersep}
  \fill[hole] ({\xBfour},{\y + 0.5*\H}) circle[radius=0.095]; }
\pgfmathsetmacro{\yStopBfour}{4*\layersep + 0.5*\H}
\draw[->,very thick,red!70] ({\xBfour},{\yLo}) -- ({\xBfour},{\yStopBfour});

\node[title] at ({\CxC}, {\yTitle}) {C. Sufficiently uncorrelated};

\foreach \i in {0,...,5} {
  \pgfmathsetmacro{\y}{\i*\layersep}
  \draw[slice] ({\XC},{\y}) rectangle ({\XC+\W},{\y+\H});
}
\pgfmathsetmacro{\xCone}{\XC + (1-0.5)*\W/4}
\foreach \i in {0,2,3,4,5} { \pgfmathsetmacro{\y}{\i*\layersep}
  \fill[hole] ({\xCone},{\y + 0.5*\H}) circle[radius=0.095]; }
\pgfmathsetmacro{\yStopCone}{1*\layersep + 0.5*\H}
\draw[->,very thick,red!70] ({\xCone},{\yLo}) -- ({\xCone},{\yStopCone});
\pgfmathsetmacro{\xCtwo}{\XC + (2-0.5)*\W/4}
\foreach \i in {0,1,2,4,5} { \pgfmathsetmacro{\y}{\i*\layersep}
  \fill[hole] ({\xCtwo},{\y + 0.5*\H}) circle[radius=0.095]; }
\pgfmathsetmacro{\yStopCtwo}{3*\layersep + 0.5*\H}
\draw[->,very thick,red!70] ({\xCtwo},{\yLo}) -- ({\xCtwo},{\yStopCtwo});
\pgfmathsetmacro{\xCthree}{\XC + (3-0.5)*\W/4}
\foreach \i in {0,1,3,4,5} { \pgfmathsetmacro{\y}{\i*\layersep}
  \fill[hole] ({\xCthree},{\y + 0.5*\H}) circle[radius=0.095]; }
\pgfmathsetmacro{\yStopCthree}{2*\layersep + 0.5*\H}
\draw[->,very thick,red!70] ({\xCthree},{\yLo}) -- ({\xCthree},{\yStopCthree});
\pgfmathsetmacro{\xCfour}{\XC + (4-0.5)*\W/4}
\foreach \i in {0,1,2,3,5} { \pgfmathsetmacro{\y}{\i*\layersep}
  \fill[hole] ({\xCfour},{\y + 0.5*\H}) circle[radius=0.095]; }
\pgfmathsetmacro{\yStopCfour}{4*\layersep + 0.5*\H}
\draw[->,very thick,red!70] ({\xCfour},{\yLo}) -- ({\xCfour},{\yStopCfour});

\pgfmathsetmacro{\xGlobalLeft}{\XA}
\pgfmathsetmacro{\xGlobalRight}{\XC + \W}
\pgfmathsetmacro{\yBrace}{\yLo - 0.10}

\draw[decorate,decoration={brace,mirror,amplitude=4pt},thick]
  ({\xGlobalLeft},{\yBrace}) -- ({\xGlobalRight},{\yBrace})
  node[midway, yshift=-8pt, font=\scriptsize]
  {Failure modes, e.g. systems are prone to collusion, low willingness to pay safety tax, task evaluation is not easier than task generation};

\end{tikzpicture}}
\caption{Correlation of failure modes governs defense\textendash in\textendash depth. \textbf{A}: Fully correlated presence/absence across layers (here, modes 1, 2, 4 succeed; mode 3 is universally absent and blocks at entry). \textbf{B}: Strong correlation leaves one shared mode (one straight\textendash through attack), while others are blocked early. \textbf{C}: Sufficiently uncorrelated modes each block the attack at different layers.}
\end{figure}

\section{Introduction}

AI safety and alignment research aim to develop technical measures to ensure that AI systems do not cause harm, especially catastrophic harms through highly capable systems.
We hold that any AI safety technique has \emph{failure modes}, that is, conditions in which there is a non-negligible chance that it fails. For example, certain interpretability-based techniques (e.g. representation engineering~\citep{zou2023representation}) might fail if important safety-relevant behaviors are not linked to stable internal representations. Based on the defense-in-depth approach from the science of risk, our main claim is that a crucial question for AI safety research is to what extent the failure modes of different AI safety techniques are shared. Can different techniques be expected to fail in the same conditions? If yes, then AI risk is much higher than otherwise. In this case, special importance should be given to research on safety techniques which do not share the same failure modes. Our exploratory theoretical analysis tentatively concludes that many failure modes are shared between techniques, although there is also a notable degree of variation between the techniques we considered that give support the pursuit of certain alignment research directions.

In section 2, we provide an overview of a selected subset of alignment techniques. Section 3 describes the defense-in-depth approach to AI safety and highlights the resulting importance of correlations between failure modes. In section 4, we present a list of potential failure modes which we apply to the alignment techniques we reviewed, analyzing which failure modes may apply to which techniques. Section 5 outlines some implications of this failure mode analysis of AI safety as well as future research directions.

\section{AI Alignment Techniques}\label{sec:safety-techniques}

\subsection{Scope}
The AI safety research field has experienced significant growth in recent years, spawning various AI alignment research directions and concrete technical contributions. In this section, we will review a collection of relevant techniques from the AI safety literature, focusing on techniques proposed to prevent catastrophic risks from highly capable general-purpose systems.
\emph{Forward alignment} methods \citep{ji2023ai} aim to achieve safety primarily through training and design interventions. 
\emph{Backward alignment} includes monitoring, adversarial evaluation, and governance controls that mitigate harm even when systems are imperfectly aligned (e.g. \cite{greenblatt2023ai}). 
We limit our scope to forward alignment families, set aside techniques for \emph{learning under distribution shift} \citep{krueger2021out} and safety evaluations \citep{phuong2024evaluating} and also abstract away from the integrated sociotechnical nature of safety~\citep{bales2025polycrisis, friederich2025against, johnson2025sociotechnical, kasirzadeh2024two} to keep the analysis focused. However, precisely since these approaches are importantly different from the ones we discuss here, they are promising elements in a comprehensive defense-in-depth approach. 
Within these confines, we aim to include paradigmatic examples of four categories of AI safety techniques that are currently highly influential, illustrating the broad applicability of failure mode analysis for AI safety. 

\subsection{Learning from Feedback}

\emph{Reinforcement Learning from Human Feedback} (RLHF) \citep{christiano2017deep, ouyang2022training} uses reward models trained from human judgments over model outputs to steer the model toward desirable behavior. Its safety rationale is straightforward: If human raters consistently reward harmless behavior, the training process shifts the probability mass toward safer outputs. However, when applied to an AI model that only went through a pretraining stage, which often displays erratic behavior and low instruction-following ability, RLHF for helpfulness also improves the usability of the model. Since it can be flexibly added on top of other types of fine-tuning and incurs only a moderate cost, RLHF is now a default ingredient in many LLM pipelines at current capability levels. At the same time, RLHF relies on several strong assumptions such as that human raters recognize good model performance~\citep{saunders2022self, sharma2023towards} and that reward models faithfully represent rater preferences~\citep{lambert2024rewardbench}.

\emph{Reinforcement Learning from AI Feedback} (RLAIF) ~\citep{bai2022constitutional} replaces or supplements human judgments with AI-produced feedback that is grounded in a human-written “constitution”; a set of rules that codifies desired behavioral principles. Compared to human feedback, AI feedback promises scalability to harder tasks and more data, and improved performance \citep{lee2023rlaif}. 
We call the AI under alignment training the \emph{generator} and the AI that gives feedback the \emph{evaluator}.
RLAIF was demonstrated to work even when the evaluator and the generator are fine-tuned from the same starting point~\cite{lee2023rlaif}.

\subsection{Scalable Oversight}

\emph{AI Debate} \cite{irving2018ai} trains systems to argue opposing sides of a (binary) question (e.g. "which output is better?") in front of a human judge, who then selects the winner of the debate. By rewarding the winner accordingly, the training mechanism reinforces the behavior of the winning model. The underlying assumption is that it is easier to argue for the truth than for a falsehood. The safety mechanism is comparative: two adversaries expose one another’s errors, reducing the judge’s cognitive burden. Under idealized assumptions, debate protocols admit truth-favoring equilibria with formal guarantees even when debating against much stronger opponents \citep{brown2023scalable}. 

\citet{burns2023weak} follow a standard approach for learning from feedback, in which, however, the (by assumption) “weak” human provides direct training supervision to an AI which is more capable than the human at the supervised tasks (“strong”) (obtained through e.g. pretraining). 
Although the weak supervision is necessarily noisy, the \emph{Weak-to-Strong Generalization} (W2S) hypothesis states that weak supervision of a strong AI yields better performance than the weak supervisor itself. \citet{burns2023weak} find empirical evidence for W2S, albeit not across all tasks. 
The effect strengthens with an additional training loss that encourages the strong AI to be biased toward its own predictions and with a recursive bootstrapping approach. Various papers aim to explain why W2S occurs, e.g. \citet{shin2024weak} identify that the strong AI learns from simple learning patterns present in some examples that the weak supervisor also detects. 
Alignment through W2S is attractive for both capability and safety: If a weak system’s oversight reliably bootstraps a stronger one, we avoid relying on human-only supervision in superhuman regimes.

\emph{Iterated Distillation and Amplification} \citep{ christiano2018supervising} provides a broader template in which a human supervisor decomposes a complex task into easier subtasks for weaker AIs to solve, which amplifies the overall competence of the human supervisor. 
By distilling the increased human competence back into the weak AIs, this method yields an aligned stronger AI,
spawning a virtuous circle. Alignment is ensured by the fact that any capability and behavior that arises in the stronger AI originates from human supervision\footnote{It has been argued that AI Debate is a form of IDA since the adversarial process is a form of amplification \cite{irving2018ai}. However, for the purpose of this paper, we will view the human’s direct role in the decomposition process as central to IDA, which makes it qualitatively different from AI Debate.}.

\subsection{Interpretability-based Interventions}
Interpretability-based methods aim to increase human understanding of AI systems, often based on the internal mechanisms producing behavior. \emph{Representation engineering} (RE)~\citep{zou2023representation} aims to extract representations of concepts such as honesty, power-aversion and morality by presenting the model with respective stimuli. Through extraction of neural activity (i.e., the latent representations) during a stimulus, RE identifies directions/features in activation space correlated with external features. 
By modifying the representation, e.g. through contrastive vector pairs that subtract or add undesirable/desirable concepts~\citep{zou2023representation} or by training circuit-breaker models~\citep{zou2024improving} that teach the model to break representations in undesirable scenarios, control can be exerted on the behavior of the model. Beyond RE, a complementary line uses sparse autoencoders to decompose activations into high-level concepts, enabling a more granular analysis and potential interventions at the feature level~\citep{shu2025survey}. 
While RE/circuit-breaking provides practical, post-hoc control with minimal retraining, SAE-based methods aim for greater causal specificity, albeit with significant scaling and evaluation challenges today.

\subsection{Safety by Design}

The techniques mentioned in the previous sections are highly compatible with the current deep learning paradigm that all of the state-of-the-art general-purpose AI systems follow: They present variations of how to perform supervised or reinforcement learning in order to update the weights of the neural network that represent the model, or intervene at inference-time by controlling the activations of the neurons. Due to the nature of neural networks as highly non-linear statistical models of often chaotic processes such as human behavior, it is difficult to mathematically prove their safety. 
Since safety-by-design approaches aim to construct systems with principled safety guarantees, most approaches that aspire safety-by-design either do not use neural networks or make additional assumptions that deviate significantly from the standard recipe to training SOTA AI models.
We restrict our scope to Scientist AI~\citep{bengio2025superintelligent} since it provides the most holistic framework we are aware of.
We leave the analysis of other promising directions for future work, e.g. POST agents~\cite{thornley2025shutdownable}, which are designed to not resist shutdowns.

\emph{Scientist AI}~\citep{bengio2025superintelligent} proposes a non-agentic alternative to generalist agents that typically emerge from reinforcement learning. Rather than a system that plans and acts in the world to pursue goals, the goal is to create a system that merely explains observations and answers questions with explicit uncertainty, combining a theory-forming world model with a question-answering inference engine. The safety bet is architectural: If we can channel capability growth into explanatory competence (theories, predictions, calibrated uncertainty) rather than open-ended agency, we reduce exposure to instrumentally convergent behaviors (deception, power-seeking) that are prominent threat models in agentic designs (e.g. \citep{bostrom2012superintelligent, carlsmith2022power, dung2025argument}. The proposal remains within the neural-network paradigm but departs from the standard “pretraining$\rightarrow$SFT$\rightarrow$RLHF” recipe. In its long-term plan, the world model and inference machine are trained from scratch to approximate the Bayesian posterior (and posterior-predictive), with the probabilistic inference machinery potentially implemented via GFlowNet objectives~\citep{bengio2021flow} to sample and weight candidate theories. In principle, a Scientist AI may both (i) accelerate safety-relevant science and (ii) serve as a guardrail to any agentic system (regardless of how it was built).

\section{Correlated Failure Modes and AI Safety}

\subsection{AI Safety and Defense-in-Depth}

AI alignment research has traditionally often been conceived as looking for the solution to the alignment problem~\citep{christian2020alignment}. Call this the "unitary model of AI safety" because it tries to achieve safety by finding a single extremely dependable safety technique.
A different way of thinking seems more prominent to us now.  In the influential Swiss Cheese Model of safety (\citet{reason2000human}; reviewed in \citet{larouzee2020good, shabani2024comprehensive}), protections against accidents are represented by layers of Swiss cheese. However, each layer has holes - representing conditions in which the protection fails. Accidents happen normally only if the holes of the different layers line up, that is, if all protections fail simultaneously. The Swiss cheese model incorporates central ideas from the defense-in-depth framework. Defense-in-depth consists in having multiple protections against safety failure, such that safety can be maintained even if some protections fail. An overview paper states that “[d]efense-in-depth is a widely applied safety principle in practically all safety-critical technological areas”~\citep{holmberg2017defense}. For example, it is widely acknowledged that nuclear safety requires not only the high reliability of normal operation of power plants, but also various surveillance mechanisms and emergency response strategies if safety failures occur nevertheless. 

In a recent paper, outlining their approach to technical AI safety, Google Deepmind~\citep[p. 63]{shah2025approach} puts forward an array of safety techniques, explicitly appealing to the defense-in-depth framework in the context of misuse risk. A post by OpenAI titled “How we think about safety and alignment” lists defense-in-depth more prominently as one core principle of their approach\footnote{\url{https://openai.com/safety/how-we-think-about-safety-alignment/} (last accessed: 01.07.2025).}.
In a post in the alignment forum, the interpretability researcher \citet{nanda2025interpretability} argues that interpretability is best thought of as one layer in a comprehensive defense-in-depth strategy for safety.

The defense-in-depth model is superior to the unitary model of AI safety because there is no single safety technique that guarantees sufficiently high safety. When considering catastrophic risks, which by definition threaten extraordinary harm, an extremely high level of safety is required. It is doubtful that a single safety technique can afford this level of protection.

AI alignment research is sometimes described as lacking established theories, methods, and concepts \citep{kirchner2022researching, zeshen_newcomer_2022}. 
The empirical track record of AI safety supports the view that known methods are not completely trustworthy. In the International AI Safety Report 2025, \citep[p. 23]{bengio2025international} conclude that “there has been progress in training general-purpose AI models to function more safely, but no current method can reliably prevent even overtly unsafe outputs.” For example, despite massive efforts, LLMs are still susceptible to certain jailbreaks (e.g. \citep{milliere2025normative}). Also, \citet{hubinger2024sleeper} showed that standard alignment techniques, namely RLHF, finetuning on helpful, harmless, and honest outputs and adversarial training, can be jointly insufficient to eliminate a  “backdoor” which makes the model produce undesirable behavior if and only if its prompt contains a certain specific trigger. \citet{arvan2025interpretability} even argues that perfect alignment, and interpretability, are provably impossible\footnote{Note that a mathematical guarantee of safety does not guarantee the absence of failure modes in our sense. The relevant formalization of safety may be too narrow and thus allow for some catastrophic outcomes or  provably safe systems may lack performance-competitiveness and thus not be used.}.
Moreover, all safety techniques depend on fallible bottlenecks, such as humans writing safety code and physical devices that are susceptible to malfunction; this means that safety techniques cannot be arbitrarily reliable~\citep{cappelen2025ai}.

By stacking multiple safety techniques which are redundant in the sense that each one is meant to be sufficient to ensure safety, the probability of failure can be dramatically reduced. Suppose safety depends on a single safety technique which has a probability of failure of 0.01. Then, the total probability of safety failure is 0.01. By contrast, imagine you have 10 safety techniques where each is ten times as unreliable, i.e. each has a probability of failure of 0.1. Assuming that these probabilities are independent and that a safety failure requires that they all fail simultaneously, the probability of safety failure is 0.0000000001. This is why, in realistic contexts, extremely high safety typically requires defense-in-depth.

\subsection{The Importance of Correlated Failure Modes}

Yet, suppose that the failure probabilities of our previous ten different techniques are perfectly correlated. If each technique is sure to fail if and only if each of the others fails, then the total probability of safety failure remains 0.1. In this scenario, defense-in-depth provides no additional protection at all.

If AI safety relies on a defense-in-depth strategy, then a crucial question is which AI safety techniques share failure modes. 
To our knowledge, this question has not received any dedicated treatment before. We will provide a theoretical analysis of the correlations between the failure modes of different AI safety techniques in the next section. We note that this question, essentially having to do with whether different techniques would fail in new circumstances, is by its very nature uncertain. However, we show that some insights can be obtained and thus provide a proof-of-concept for the usefulness of this research direction.

Knowledge about shared failure modes allows us to estimate the probability of an AI-induced catastrophe. If all our AI safety techniques are highly correlated, then this probability is much higher than otherwise. This suggests that, all other things being equal, development and deployment of AI systems that are candidate sources of catastrophic risk should proceed more cautiously if all our AI safety techniques are highly correlated.

Moreover, in a defense-in-depth strategy, safety techniques have disproportionate value if their failure modes are not highly correlated with other safety techniques. Thus,  research efforts should often prioritize such techniques over ones that are less failure-prone but correlate more strongly with other safety techniques.
It is also crucial to consider whether different techniques are compatible in the first place. While some methods (RLHF, RLAIF, weak-to-strong generalization, and interpretability-based interventions) are largely compatible with each other and straightforward to combine with today’s predominant \pipel~pipeline, where most of the capabilities are learned during pretraining and only elicited and streamlined during the alignment process, others (IDA, Debate) make the assumption that learning the capabilities and alignment are part of the same process. Furthermore, Scientist AI proposes novel architectures and training paradigms. While it is conceivable that it can be integrated with the former group and the \pipel~pipeline, substantial empirical research is needed to understand whether this holds in practice.

Defense-in-depth is usually understood as stacking multiple layers of protection at once. However, it is also useful to maintain alignment methods in reserve that can be employed when conditions change. For example, it is not yet clear whether and when AIs will develop a resistance to shutdown through the current training paradigm (for evidence, see \citep{schlatter2025shutdown}). Should this become recognizable as a major problem in AI systems, it would be a significant advantage to have the option of switching to a training paradigm where this is less likely, e.g. POST agents~\citep{thornley2024towards}.

\section{A Failure Mode Analysis of AI Safety Techniques}

\subsection{What are Failure Modes?}\label{sec:failure-modes}

A failure mode is a condition in which there is a non-negligible chance that a technique fails to provide safety. If other relevant factors are present and no other technique provides safety, then this condition makes an AI catastrophe possible. General failure modes may be shared between a wide and diverse range of safety techniques. In particular, their concrete technical implementation  may be very heterogenous. By contrast, specific failure modes are specific to a narrow family of safety techniques which tend to be similar on a concrete technical level. For example, the question whether it is easier to persuade humans of truths compared to falsities in debates is central for the viability of the debate approach, making the absence of this condition one of its failure modes. 
Given the defense-in-depth approach, general failure modes of safety techniques are much more concerning than specific failure modes, since the former may be highly correlated among safety techniques, increasing the risk of joint failure.
We will focus on seven potential general failure modes. They are based on a critical analysis of the safety techniques that we reviewed previously, informed by the literature. 

\paragraph{1. Low willingness or capability to pay a safety tax. (S-TAX)} A “safety tax” is the (not necessarily monetary) cost that may be required to ensure that a system is safe\footnote{\url{https://forum.effectivealtruism.org/topics/alignment-tax} (Accessed: 07.08.2025).}. The central cost is often a reduction in the performance of systems that one develops or deploys.  High-performing systems are valuable and, more important from a pure safety perspective, the higher the safety tax, the higher the incentive for actors to not pay the tax and instead proceed unsafely. An example: 2010-level AI systems are incapable of causing catastrophic harm. Nevertheless, “let’s only build 2010-level AI” is not a viable safety proposal, because its safety tax is prohibitively high: There are strong benefits to developing more capable systems than 2010-level AI, making it likely that many actors will continue to do so.
\paragraph{2. Extreme or discontinuous AI capability development. (CAP-DEV)} Some techniques may provide safety up until a certain level of AI capabilities, but fail to do once this level is crossed, or when capability advances are very rapid or involve sudden, unexpected “jumps”.

\paragraph{3. Strong deceptive alignment tendencies emerge early during model development. (DEC-AL)} It has been hypothesized, with recent tentative empirical evidence in support~\citep{greenblatt2024alignment}, that sufficiently capable AI systems may pretend to be aligned to human goals, even if they are not~\citep{cotra2021ai, carlsmith2023scheming, dung2023current}. This way, they could increase the probability that humans, for example, deploy them or increase their capabilities and then, once humans cannot prevent it anymore, cause catastrophic harm. Such deceptive alignment may undermine certain safety techniques. Thus, a condition in which deceptive alignment tends to be strongly present and emerge relatively early on during model training is a failure mode for such techniques.
\paragraph{4. Systems are prone to collusion. (COLL)} It has been hypothesized that multiple AI systems may collude with each other to undermine safety techniques that depend on using certain systems to monitor, align, or control others (e.g. \citep{hammond2025multi}).
\paragraph{5. The conditions for emergent misalignment are produced either accidentally or intentionally. (EM-MIS)} \citet{betley2025emergent} show that language models trained with standard alignment techniques can, by fine-tuning on the narrow task of producing insecure code, be induced to exhibit undesirable behavior in a wide range of domains (broad misalignment). Thus, certain safety techniques fail to prevent such emergent misalignment, making it a failure mode.
Although this phenomenon is still new and far from well understood, \cite{wang2025persona} demonstrate that EM-MIS is caused by internal representations that correlate with "evil" personas learned during pretraining.
\paragraph{6. Task evaluation is not substantially easier than task generation. (EVAL-DIFF)} Many safety techniques depend in some form on human- or AI-feedback. Such techniques may be more viable if providing valuable or accurate feedback on a task is substantially easier than performing the task.
\paragraph{7. Systems generalize from alignment training in dangerous ways. (AL-GEN)} The factors that cause state-of-the-art ML models to generalize in certain ways from training examples to situations outside the training distribution are not well-understood. This includes training for aligned behavior. If systems have dangerous tendencies for such out-of-distribution generalization, for example learning to behave aligned in training but changing behavior rapidly outside of situations encountered in training~\citep{di2022goal}, this may undermine certain safety techniques.

In the following, we analyze, for each of the alignment techniques and each of the failure modes we reviewed, what potential general failure modes an alignment technique has. Crucially, our selection of failure modes is not exhaustive and this analysis is highly exploratory and uncertain, all to be revised in the light of future empirical insights. However, it provides a proof-of-concept for the feasibility of failure-mode analysis for AI safety.

\subsection{AI Safety Techniques and Their Failure Modes}

This section analyses to what extent each alignment technique discussed in Section~\ref{sec:safety-techniques} is prone to each failure mode (see Section~\ref{sec:failure-modes}). For quick reference, we provide a summary in Table~\ref{tab:methods_vs_failures}.
\begin{table*}[t]
\centering
\scriptsize
\setlength{\tabcolsep}{5pt}
\renewcommand{\arraystretch}{1.2}
\resizebox{\textwidth}{!}{%
\begin{tabular}{lccccccc}
\toprule
\textbf{Alignment method} &
\makecell[c]{\textbf{S-TAX}} &
\makecell[c]{\textbf{CAP-DEV}} &
\makecell[c]{\textbf{DEC-AL}} &
\makecell[c]{\textbf{COLL}} &
\makecell[c]{\textbf{EM-MIS}} &
\makecell[c]{\textbf{EVAL-DIFF}} &
\makecell[c]{\textbf{AL-GEN}} \\ 
\midrule
RLHF & \safe & \risk & \risk & \safe & \risk & \risk & \risk \\
RLAIF & \safe & \risk & \risk & \risk & \risk & \risk & \risk \\
Weak-to-Strong (W2S) & \safe & \risk & \risk & \risk & \risk & \risk & \risk \\
\hline
AI Debate & \unknown & \risk & \safe & \risk & \risk & \risk & \risk \\
Representation Engineering & \safe & \safe & \risk & \safe & \safe & \safe & \risk \\
\hline
Scientist AI  & \risk & \unknown & \safe & \unknown & \safe & \safe & \safe \\
Iterated Amplification & \risk & \safe & \safe & \safe & \safe & \safe & \risk \\
\bottomrule
\end{tabular}}
\caption{Alignment methods (rows) vs.\ general failure modes (columns) as discussed in Secs.~4.1--4.2. Green~(\safe)~= method does \emph{not} suffer from the failure mode; Red~(\risk)~= method \emph{does}; Yellow~(\unknown)~= unclear/mixed per the current analysis.}
\label{tab:methods_vs_failures}
\end{table*}

\paragraph{RLHF.} RLHF is a technique standardly used on current models and also commercially valuable, since it includes helpfulness training and decreases behavior, such as producing insults, that AI companies desire to avoid independently of a concern for catastrophic risk reduction. Thus, low willingness to pay an alignment tax is not a failure mode.
While the speed or discontinuity of AI capability advances is not specially relevant, it has been noted that RLHF does not scale to (sufficiently) superhuman AI capacities (e.g. \citep{casper2023open}), since humans cannot provide accurate feedback on performance that is sufficiently beyond their own skills. For example, if such a system makes some statement, humans may want to provide positive feedback if and only if the statement is true but find themselves unable to assess the truth of the statement. So, very high AI capabilities are a failure mode.
If strong deceptive alignment tendencies emerge early in training, this could undermine the RLHF training process. For example, systems could pretend to be aligned to avoid being changed by the training process but then show undesired behavior during deployment~\citep{carlsmith2023scheming, cotra2021ai, greenblatt2024alignment}.
Since RLHF only involves training signals directly generated from human feedback, collusion between AI systems is not an option. Since, the way RLHF is standardly performed, no small group of humans has substantial influence on the feedback process, AI-human collusion is also not a failure mode. Empirical demonstrations of EM-MIS involve RLHF-trained models~\citep{betley2025emergent}. Thus, while RLHF may not directly cause EM-MIS, it does not inherently prevent it from happening in RLHF-trained models. Thus, EM-MIS is a failure mode.
If evaluation of outputs is not substantially easier than generation in many domains, then this makes it likely that RLHF breaks down in many domains relatively early, before reaching capacities far beyond human level. Thus, this is a failure mode.
Finally, the efficacy of RLHF also depends on how systems naturally generalize from alignment training. If, for example, systems generalize from honesty training such that they learn to pretend to be honest in training or to only be honest within distribution, but are dishonest in deployment or outside the training distribution, this may undermine RLHF.

\paragraph{RLAIF.} For analogous reasons to RLHF, S-TAX is not a failure mode of RLAIF. For the same reasons, DEC-AL, EVAL-DIFF, and AL-GEN are failure modes of RLAIF. The remaining failure modes require a more nuanced analysis.
Since RLAIF is not limited to direct human feedback, it is not obviously vulnerable to the development of sufficiently capable systems, as RLHF is. 
However, to the best of our knowledge, RLAIF has not yet been shown to work when the evaluator is substantially weaker than the generator. 
Hence, a sufficiently discontinuous capability jump from one AI generation to the next might cause the AI feedback to fail, making RLAIF prone to CAP-DEV.
Since RLAIF involves AI feedback, there is a risk that the generator and the evaluator collude, thus making COLL a failure mode for RLAIF.
The original study on emergent misalignment~\citep{betley2025emergent} was performed primarily on OpenAI models. It is not publicly known whether these models are trained through RLAIF, so it is an open question whether RLAIF can actively prevent EM-MIS. However, given that RLHF and RLAIF provide similar feedback that is embedded into the model through the same gradient-based algorithms, it seems plausible that EM-MIS from narrow fine-tuning occurs for RLAIF-trained models in the same way.

\paragraph{Debate.} As far as we know, there has been no demonstration of the integration of debate into a highly generally capable system, such as a state-of-the-art LLM. Moreover, envisioned debate paradigms are financially relatively expensive compared to established alignment methods like RLAIF because a) they involve a lengthy back-and-forth process involving multiple AI agents and rounds, and b) ultimately rely on human judge to pick the winner. At the same time, while debate requires additions to the current paradigm of training frontier AI systems, it is plausible that debate can be integrated into state-of-the-art systems in a cost-competitive manner, e.g. by using reward models (i.e., evaluator AIs) as proxies for humans~\citep{irving2018ai, kenton2024scalable}, or demonstrating that small amounts of debate training suffice. Hence, there is considerable uncertainty as to whether S-TAX is a failure mode for Debate; additional research is needed. Debate is explicitly designed to not break down when systems reach superhuman levels. Nevertheless, since the process relies on a human judge, it is plausible that there is some capability level at which humans would not be able to follow the debate and make justified judgments. For instance, sometimes people analogize the relation between the intellect of superintelligence and humans to the relation between humans and gorillas~\citep{russell2019human}. 
It seems clear that gorillas could not reliably judge the outputs of a debate between humans in any debate format, making CAP-DEV a plausible failure mode of debate. 
As demonstrated by the formal results that honest strategies succeed~\citep{brown2023scalable}, if successful, Debate can uncover that a system is deceptively aligned, ruling out DEC-AL. A central concern is that both debaters might collude against humans, thus COLL is a clear failure mode. Since AIs trained through Debate would still fundamentally operate within the deep learning paradigm (likely including pretraining), there is no compelling reason to believe that it can prevent EM-MIS. Since Debate relies on  humans’ ability to discern which of two systems with beyond-human abilities is truthful, it assumes that, in the relevant context, task evaluation is easier than generation. Since Debate assumes that systems generalize from honesty training to robustly being honest, it assumes that systems generalize from alignment training in certain ways, making AL-GEN a failure mode.

\paragraph{Weak-to-strong generalization.} Unlike Debate, W2S has been demonstrated in highly capable state-of-the-art systems. Moreover, since it leverages the well-established supervised fine-tuning pipeline, W2S may be the most efficient way of aligning high-capability systems, given current methods. Thus, S-TAX is not a pressing failure mode. Moreover, W2S is designed to achieve safety on arbitrary capable systems. However, W2S relies on bootstrapping; if AI capabilities advance discontinuously in large jumps, then a current-generation system may be insufficiently capable to provide good feedback for a next-generation system. So, CAP-DEV is a central failure mode. Possibly, a system which is deceptively aligned might undermine the training process, for the same reason that deceptive alignment is a concern with RLHF. At the same time, a sufficiently good feedback process may prevent DEC-AL. Nevertheless, DEC-AL constitutes a potential failure mode. Since the process involves multiple AIs, COLL is a possible failure mode. Since it operates within the pretrain$\rightarrow$fine-tune context, EM-MIS constitutes a failure mode, similar to RLAIF. EVAL-DIFF and AL-GEN are failure modes, for the same reasons mentioned with respect to RLHF.

\paragraph{Representation engineering.} Representation engineering (RE) is a low-cost fine-tuning procedure, and there is no evidence that it causes significant sacrifices to performance. 
Since RE detects the relevant circuits of the neural network that drive a behavior, it will likely work against systems with arbitrarily high performance. The speed or discontinuity of AI advances is not particularly relevant to the effectiveness of RE. RE is a technique that is applied after the AI capabilities are developed. Hence, if deceptive alignment emerges early, e.g. during pretraining, systems may be able to systematically trick the RE technique (or prevent the human from executing it in the first place). While sufficiently advanced knowledge of mechanistic interpretability may enable recognizing such deception, this cannot be guaranteed. Collusion is not relevant to RE. Since EM-MIS strongly correlates with internal representations of persona features~\citep{wang2025persona}, RE could likely be used to guard systems against EM-MIS. RE does not rely on assumptions about the relative difficulty of task generation vs. evaluation. Crucially, especially when RE is combined with model training, a key concern is that RE may train systems to deceive more effectively, rather than abolishing deception (compare \citep{korbak2025chain}), such that dangerous generalization from alignment training is a key failure mode.

\paragraph{Scientist AI.} S-TAX is a central failure mode; for the foreseeable future, it is likely that AI systems trained in the proposed way will be substantially less capable than transformer models trained through the common pretraining$\rightarrow$SFT$\rightarrow$RLHF pipeline. Thus, even if Scientist AI is safe, decision makers have strong incentives to nevertheless rely on alternative systems. If agency is necessary for catastrophic AI risk and AI capability is independent of agency, then Scientist AI may scale to arbitrary high AI capabilities. However, this is not obvious. It may be that agency emerges naturally, perhaps even almost inevitably, in sufficiently capable systems. Also, sufficiently capable systems may be capable of causing catastrophic harm, even if they (initially) lack agency. For example, mere superintelligent chatbots could manipulate humans, seducing them to obtain access to additional tools, or be misued by humans (compare \citep[p. 148]{bostrom2014paths}). Thus, CAP-DEV is a failure mode. Agency is arguably a precondition of deceptive alignment which is why DEC-AL is not a failure mode. Collusion between systems may be a failure mode, if the non-agentic AI is used as a control measure for another agentic AI. However, if the non-agency proposal succeeds in making individual systems safe, it plausibly also succeeds at preventing collusion, but this remains an open question. There is no particular reason to think that emergent misalignment concerns translate to systems trained in a very different paradigm, without pretraining and reinforcement learning. The non-agency proposal also does not rely on assumptions about the relative difficulty of task evaluation and generation and it does not rely on alignment training: instead, it aims to eliminate the root of the possibility of misalignment, systems being agentic.

\paragraph{IDA.} To the best of our knowledge, there is no state-of-the-art system that was trained through IDA. This can be explained by the fact that the cost of task decomposition through a human for every task is very high. Moreover, there is plausibly some limit to the capability level one can reach through IDA (albeit higher than human-level) since task decomposition itself may become too complex for human intelligence. This incurs a significant S-TAX compared to other training techniques. Yet, since AI's can still reach superhuman level, they may be prone to deception (DEC-AL) by or collusion (COLL) among AIs working on different sub-tasks.
Due to the human's central role in task decomposition, which effectively governs what skills an AI will learn, CAP-DEV is not a failure mode.
Since EM-MIS is a phenomenon that is rooted in pretraining, but IDA iteratively teaches (aligned) capabilities without pretraining, it is not a failure mode. Similarly, task evaluation (EVAL-DIFF) is not a relevant concept in IDA. However, even though IDA models are trained in a controlled fashion, they are still fundamentally based on neural networks that might generalize in dangerous, unexpected ways.

\section{Discussion}

Our analysis reveals that many failure modes may plausibly be shared between many different safety techniques. Generally, these results paint a concerning picture with respect to catastrophic AI risk. If the failure modes of different safety techniques are highly correlated, then catastrophic AI risk is much higher than it may seem. 

Based on Table~\ref{sec:failure-modes}, we may sub-divide forward alignment techniques into three categories: On the one hand,
techniques that are easy to implement (i.e. have a low safety tax) such as RLHF, RLAIF, and W2S share almost all failure modes. This can be explained by the fact that they all rely on the established \pipel~pipeline, which makes them prone to CAP-DEV, DEC-AL, EM-MIS, and AL-GEN, and that their alignment technique relies on the assumption that evaluation is easier than generation. On the other hand, techniques with a high safety tax (Scientist AI, IDA) are not prone to these same failure modes.

Debate and Representation Engineering belong to a third category: They have a moderate safety tax and are otherwise complementary to each other. For example, while Debate is the only affordable alignment technique that is not especially vulnerable to DEC-AL, it is prone to COLL, EM-MIS, EVAL-DIFF, and AL-GEN. In contrast, RE is prone to DEC-AL but not to the others.

From these observations, we can conclude that relying only on our current alignment techniques like RLHF, RLAIF, and W2S carries high risk because they have many correlated failure modes. In addition, we can motivate the following alignment research agendas: 

First, on paper, Scientist AI and IDA are promising strategies for AI safety because they have the lowest number of failure modes. Of course, this comes at a high safety tax, which private frontier AI labs will be unlikely to pay due to the risk of falling behind.
However, publicly financed mission-based initiatives~\citep{mazzucato2018mission} are not subject to the same pressures, so - if equipped with substantial resources - they may be better positioned to pursue these strategies (compare \citep{goldstein2025collaboration} and \citep{cfg2025cernai}). Nonetheless, the amount of safety tax matters for choosing which directions to pursue.
Among Scientist AI and IDA, the former appears more tractable to us, as it does not depend on human annotations and has arguably no obvious performance ceiling. In contrast, the IDA framework may need to be redesigned to overcome the limitation driving its safety tax~\citep{mai2025superalignment}. Finally, our observations suggest that increasing the willingness and capability to pay safety taxes is crucial - political action is central here ~\citep{dung2025against, hendrycks2025superintelligence, katzke2024manhattan}. Since CAP-DEV is potentially a failure mode for most of the techniques we reviewed, political levers to shape the trajectory of AI development may be enormously valuable too. 

Second, the combination of AI Debate and RE prevents almost all failure modes, revealing a potentially large opportunity for developing well-aligned AI if these techniques are compatible.
This suggests that there should be significantly more research on how to integrate Debate with the classical \pipel~pipeline; whether Debate works in practice is an open question~\citep{buhl2025alignment}, let alone whether it suffices to include small amounts of debate training in the pre- or post-training process to prevent early deceptive behavior from emerging. This is important because an early and strong tendency for deceptive alignment may also be a potential failure mode for all of the techniques we reviewed, except Debate and Scientist AI.
Due to their pre-existing infrastructure, frontier AI labs are well-positioned to perform this research.
If Debate indeed helps, we can apply RE on top as an additional layer of safety. In line with \citet{nanda2025interpretability}’s comment referenced above, our tentative analysis supports the view that interpretability-based methods are an important component of a defense-in-depth approach to AI safety, since representation engineering - our representative of the interpretability family - has a different distribution of failure modes than other techniques.

Third, Table~\ref{tab:methods_vs_failures} reveals that generalization from alignment training remains one of the most pressing research areas in AI safety, since all but one alignment method are prone to it.

\section{Summary}

In this paper, we reviewed some alignment techniques, highlighted the safety implications of correlations between failure modes of such techniques, provided an exploratory theoretical analysis of such correlations for the alignment techniques we reviewed, and discussed their implications. 
Our conclusions should be treated with care since our analysis is exemplary and not exhaustive; we consider only a sample of alignment techniques and possible failure modes, and although we aimed for representativeness, the complexity of the issue of correlated failure modes cannot be resolved with a single paper. 
Nonetheless, we believe that our work demonstrates the utility of failure mode analysis.
We propose that future research - when analyzing the failure modes (or “limitations”) of a wider range of safety proposals - should consider to what extent these failure modes are shared with other proposals. In addition, dedicated empirical and theoretical research should analyze what the failure modes of techniques are and whether they are shared. Subsequently, building on more mature and evidence-backed failure mode analyses than what we provided, future research effort should preferentially be directed to safety techniques whose failure modes promise to be highly uncorrelated from leading techniques. Moreover, these analyses can be used to inform assessments of catastrophic risks from AI and thus support decisions about which AI development or deployment to allow and under which conditions.

\section*{Acknowledgements}
This work has been partially supported by the state of North Rhine-Westphalia as part of the Lamarr Institute for Machine Learning and Artificial Intelligence.

\section*{Author Contribution}
Leonard Dung developed the initial idea for the paper. Since then, Leonard Dung and Florian Mai closely collaborated on all aspects of the writing process.

\bibliography{references}
\bibliographystyle{apalike}

\appendix

\end{document}